%% file: main.tex
\documentclass[10pt,twocolumn,letterpaper]{article}

\usepackage{wacv}
\usepackage{times}
\usepackage{epsfig}
\usepackage{graphicx}
\usepackage{amsmath}
\usepackage{amssymb}
\usepackage{booktabs}

\makeatletter
\@namedef{ver@everyshi.sty}{}
\makeatother

\usepackage[accsupp]{axessibility}  

\def\wacvPaperID{1} 

\wacvapplicationstrack 

\wacvfinalcopy 

\include{preamble}
\input{acronyms}

\input{symbols}
\graphicspath{{Images/}}

\def\codelink{\url{https://github.com/jspenmar/monodepth_benchmark}}

\def\teamone{OPDAI\xspace}
\def\teamtwo{z.suri\xspace}
\def\teamthree{Anonymous\xspace}
\def\teamfour{MonoViT\xspace}

\ifwacvfinal
\usepackage[breaklinks=true,bookmarks=false]{hyperref}
\else
\usepackage[pagebackref=true,breaklinks=true,colorlinks,bookmarks=false]{hyperref}
\fi

\begin{document}

\title{The Monocular Depth Estimation Challenge}

\author{
Jaime Spencer$^{1}$ \quad
C.\ Stella Qian$^{2}$ \quad
Chris Russell$^{3}$ \quad
Simon Hadfield$^{1}$ \quad
Erich Graf$^{4}$ \quad \\
Wendy Adams$^{4}$ \quad 
Andrew J.\ Schofield$^{2}$ \quad
James Elder$^{5}$ \quad
Richard Bowden$^{1}$ \quad
Heng Cong$^{6}$ \quad \\
Stefano Mattoccia$^{7}$ \quad
Matteo Poggi$^{7}$ \quad 
Zeeshan Khan Suri$^{8}$ \quad
Yang Tang$^{9}$ \quad
Fabio Tosi$^{7}$ \quad \\
Hao Wang$^{6}$ \quad 
Youmin Zhang$^{7}$ \quad
Yusheng Zhang$^{6}$ \quad
Chaoqiang Zhao$^{9}$ \quad 
\\ \\
$^{1}$University of Surrey \quad
$^{2}$Aston University \quad
$^{3}$Amazon \quad
$^{4}$University of Southampton \quad \\
$^{5}$York University \quad
$^{6}$Independent \quad
$^{7}$University of Bologna \quad
$^{8}$DENSO ADAS Engineering  \\ Services GmbH \quad
$^{9}$East China University of Science and Technology \quad
}

\maketitle
\thispagestyle{empty}

\begin{abstract}
This paper summarizes the results of the first Monocular Depth Estimation Challenge (MDEC) organized at WACV2023. 
This challenge evaluated the progress of self-supervised monocular depth estimation on the challenging SYNS-Patches dataset. 
The challenge was organized on CodaLab and received submissions from 4 valid teams. 
Participants were provided a devkit containing updated reference implementations for 16 State-of-the-Art algorithms and 4 novel techniques.
The threshold for acceptance for novel techniques was to outperform every one of the 16 SotA baselines.
All participants outperformed the baseline in traditional metrics such as MAE or AbsRel. 
However, pointcloud reconstruction metrics were challenging to improve upon. 
We found predictions were characterized by interpolation artefacts at object boundaries and errors in relative object positioning.
We hope this challenge is a valuable contribution to the community and encourage authors to participate in future editions. 
\end{abstract}

\section{Introduction} \label{sec:intro}
Depth estimation is a core computer vision task, allowing us to recover the \ndim{3} geometry of the world. 
Whilst traditional approaches to depth estimation relied on stereo~\cite{Hirschmuller2005,Sun2003,Chang2018} or multi-view~\cite{Agarwal2011,Schoenberger2016,Lindenberger2021} matching, monocular approaches~\cite{Eigen2015,Garg2016,Godard2019,Ranftl2020} requiring only a single image have recently garnered much attention. 

The task of \ac{mde} is ill-posed, as an infinite number of scene arrangements with varying object sizes and depths could result in the same \ndim{2} image projection. 
However, humans are capable of performing this task by relying on cues and priors such as absolute/relative object sizes, elevation within the scene, texture gradients, perspective distortions, stereo/motion parallax and more. 
Networks performing \ac{mde} must also learn these geometric cues, rather than just rely on correspondence matching.

The rise in popularity of this field has resulted in a plethora of contributions, including supervised~\cite{Eigen2015,Ranftl2020}, self-supervised~\cite{Garg2016,Godard2017,Godard2019,Guizilini2020} and weakly-supervised~\cite{Rui2018,Tosi2019,Watson2019} approaches.   
Comparing these approaches in a fair and consistent manner is a highly challenging task, as it is the responsibility of each author to ensure they are following the same procedures as preceding methods. 
The need to provide this backward-compatibility can result in long-standing errors in the benchmarking procedure, ranging from incorrect metric computation and data preprocessing to incorrect ground-truths.

This paper covers the recent \ac{mdec}, organized as part of a workshop at WACV2023.
The objective of this challenge was to provide an updated and centralized benchmark to evaluate contributions in a fair and consistent manner. 
This first edition focused on self-/weakly-supervised \ac{mde}, as they have the possibility to scale to larger amounts of data and do not require expensive \acs{lidar} ground-truth. 
Despite this flexibility, the majority of published approaches train and evaluate only on automotive data.
As part of this challenge, we tested the generalization of these approaches to a wider range of scenarios, including natural, urban and indoor scenes. 
This was made possible via the recently released \acl{syns} dataset~\cite{Adams2016,Spencer2022}. 
In general, participants found it challenging to outperform the updated Garg baseline~\cite{Garg2016,Spencer2022} in pointcloud-based reconstruction metrics (F-Score), but generally improved upon traditional image-based metrics (MAE, RMSE, AbsRel).

\section{Related Work} \label{sec:lit}

%
\begin{table}[htbp]
\scriptsize
\renewcommand{\arraystretch}{1.5}
\centering
\input{Tables/*}
\label{tbl:[}
\end{table}
!t]{datasets}{lit:datasets}

To avoid using costly \ac{lidar} annotations, self-supervised approaches to \ac{mde} instead rely on the proxy task of image reconstruction via view synthesis. 
The predicted depth is combined with a known (or estimated) camera transform  to establish correspondences between adjacent images.
This means that, whilst the network can predict depth from a single input image at test time, the training procedure requires multiple support frames to perform the view synthesis. 

Methods can be categorized based on the source of these support frames. 
Stereo methods~\cite{Garg2016,Godard2017,Rui2018,Tosi2019} rely on stereo rectified images pairs with a known and fixed camera baseline. 
This allows the network to predict metric depth, but can result in occlusions artefacts if not trained carefully. 
On the other hand, monocular approaches~\cite{Zhou2017,Klodt2018,Wang2018} commonly use the previous and following frame from a monocular video.
These approaches are more flexible, as no stereo data is required.
However, they are sensitive to the presence of dynamic objects. 
Furthermore, depth is predicted only up to an unknown scale factor and requires median scaling during evaluation to align it with the ground-truth.

Garg~\cite{Garg2016} introduced the first approach to \ac{mde} via stereo view synthesis, using AlexNet~\cite{Krizhevsky2017} and an \pnorm{1} reconstruction loss.
Monodepth~\cite{Godard2017} drastically improved the performance through bilinear synthesis~\cite{Jaderberg2015} and a weighted combination of SSIM~\cite{Wang2004} and \pnorm{1}.
It additionally incorporated virtual stereo supervision and a smoothness regularization weighted by the strength of the image edges. 
3Net~\cite{Poggi2018} extended this to a trinocular setting, while DVSO~\cite{Rui2018} and MonoResMatch~\cite{Tosi2019} incorporated an additional residual refinement network. 

SfM-Learner~\cite{Zhou2017} introduced the first fully monocular framework, replacing the fixed stereo baseline with a \ac{vo} regression network. 
A predictive mask was introduced to downweigh the photometric loss at independently moving dynamic objects. 
Future methods refined this masking procedure via uncertainty estimation~\cite{Kendall2017a,Klodt2018}, object motion prediction~\cite{Lee2021,Klingner2020,Li2021} and automasking~\cite{Godard2017,Bian2019}.
Monodepth2~\cite{Godard2019} additionally proposed the minimum reprojection loss as a simple way of handling varying occlusions in a sequence of frames. 
Instead of averaging the reconstruction loss over the sequence, they proposed to take only the minimum loss across each image pixel, assuming this will select the frame with the non-occluded correspondence. 

Subsequent approaches focused on improving the robustness of the photometric loss by incorporating feature descriptors~\cite{Zhan2018,Spencer2020,Shu2020}, affine brightness changes~\cite{Yang2020a}, scale consistency~\cite{Mahjourian2018,Bian2019} or adversarial losses~\cite{Aleotti2018,Pilzer2018,Liu2021}.
Meanwhile, the architecture of the depth prediction network was improved to target higher-resolution predictions by incorporating sub-pixel convolutions~\cite{Shi2016,Pillai2019}, \ndim{3} packing blocks~\cite{Guizilini2020}, improved skip connections~\cite{Yan2021,Zhou2021,Lyu2021}, transformers~\cite{Zhao2022} and discrete disparity volumes~\cite{Johnston2020,Bello2020,Bello2021}.

Several methods incorporated additional supervision in the form of (proxy) depth regression from \ac{lidar}~\cite{Kuznietsov2017,Guizilini2021}, synthetic~\cite{Luo2018}, \acs{slam}~\cite{Klodt2018,Wang2018,Rui2018}, hand-crafted stereo~\cite{Tosi2019,Watson2019,Liu2021}, the matted Laplacian~\cite{Bello2021} and self-distillation~\cite{Pilzer2019,Petrovai2022}.
One notable example is DepthHints~\cite{Watson2019}, which combined hand-crafted disparity~\cite{Hirschmuller2005} with the min reprojection loss~\cite{Godard2019}. 
This provided a simple way of fusing multiple disparity maps into a single robust estimate.  

%
\begin{table}[htbp]
\scriptsize
\renewcommand{\arraystretch}{1.5}
\centering
\input{Tables/*}
\label{tbl:[}
\end{table}
!t]{syns_cat}{data:syns_cat}

%
\begin{figure}[htbp]
\centering
\input{Figures/*}
\label{fig:[}
\end{figure}
!t]{syns}{data:syns}

\subsection{Datasets \& Benchmarks}
This section reviews some of the most commonly used datasets and benchmarks used to evaluate \ac{mde}.
Despite being a fundamental and popular computer vision task, there has not been a standard centralized challenge such as ImageNet~\cite{Deng2009}, VOT~\cite{Kristan2022} or IMC~\cite{Jin2021}.
This makes it challenging to ensure that all methods use a consistent evaluation procedure. 
Furthermore, the lack of a withheld test set encourages overfitting due to repeated evaluation.
\tbl{lit:datasets} provides an overview of these datasets.

Kitti~\cite{Geiger2013} is perhaps the most common training and testing dataset for \ac{mde}.
It was popularized by the \ac{ke} split~\cite{Eigen2015}, containing 45k images for training and 697 for testing.
However, this benchmark contains some long-standing errors that heavily impact the accuracy of the results. 
The ground-truth depth suffers from background bleeding at object boundaries due to the different sensor viewpoints, coupled with the motion artefacts produced by the moving \ac{lidar}.
Furthermore, the data preprocessing omitted the transformation to the camera reference frame. 
These issues are further exacerbated by the sparsity of the ground-truth depth maps, which contain measurements for only 4.10\% of the image pixels.

Uhrig~\etal~\cite{Uhrig2018} aimed to correct these errors and provide a more reliable benchmark, dubbed the \ac{keb} split.
The ground-truth density was improved to 15.28\% by accumulating \ac{lidar} data from $\pm 5$ adjacent frames.
This data was aggregated and refined by adding consistency checks using a hand-crafted stereo matching algorithm~\cite{Hirschmuller2005}.
The main drawback is that this refinement procedure removes points at object boundaries, which are common sources of errors even in \ac{sota} approaches.
However, despite providing a clear improvement over \ac{ke}, adoption by the community has been slow.
We believe this to be due to the need to provide consistent comparisons against previous methods that only evaluate on \ac{ke}, as this would require authors to re-run all preexisting approaches on this new baseline. 

The DDAD dataset~\cite{Guizilini2020} contains data from multiple cities in USA and Japan and totalling to 76k training and 3k testing images. 
It provides an density of 1.03\%, an average of 24k points per image and an increased depth range up to 250 meters. 
This dataset was the focus on the DDAD challenge organized at CVPR 2021, which featured additional fine-grained performance metrics on each semantic class. 
Similar to \ac{keb}, we believe that adoption of these improved datasets is hindered by the need to re-train and re-evaluate preexisting methods. 

Spencer~\etal~\cite{Spencer2022} aimed to unify and update the training and benchmarking procedure for \ac{mde}.
This was done by providing a public repository containing modernized \ac{sota} implementations of 16 recent approaches with common robust design decisions. 
The proposed models were evaluated on the improved \ac{keb} and \acl{syns}, incorporating more informative pointclound-~\cite{Ornek2022} and edge-based~\cite{Koch2018} metrics. 
This modern benchmark procedure constitutes the basis of the \acl{mdec}.

\section{The Monocular Depth Estimation Challenge} \label{sec:meth}
The first edition of \ac{mdec} was organized as part of a WACV2023 workshop. 
The challenge was organized on CodaLab~\cite{Pavao2022} due to its popularity and flexibility, allowing for custom evaluation scripts and metrics. 
We plan to arrange a permanent leaderboard on CodaLab that remains open to allow authors to continue evaluating on \acl{syns}.

The first two weeks of the challenge constituted the development phase, where participants could submit predictions only on the validation split of \acl{syns}. 
For the remainder of the challenge, participants were free to submit to either split.
Participants only had access to the dataset images, while the ground-truth depth maps and depth boundaries were withheld to prevent overfitting. 

\subsection{Dataset} \label{sec:meth:dataset}
The evaluation for the challenge was carried out on the recently introduced \acl{syns} dataset~\cite{Spencer2022}, which is a subset of SYNS~\cite{Adams2016}.
The original SYNS is composed of aligned image and \ac{lidar} panoramas from 92 different scenes belonging to a wide variety of environments, such as Agriculture, Natural (\eg forests and fields), Residential, Industrial and Indoor.
This is a departure from the commonly used datasets in the field, such as Kitti~\cite{Geiger2013}, CityScapes~\cite{Cordts2016} or DDAD~\cite{Guizilini2020}, which focus purely on urban scenes collected by automotive vehicles. 
SYNS also provides dense \ac{lidar} maps with 78.30\% coverage and 365k points per image, which are exceptionally rare in outdoor environments. 
This allows us to compute metrics targeting complex image regions, such as thin structures and depth boundaries, which are common sources of error. 

\acl{syns} represents the subset of patches from each scene extracted at eye level at 20 degree intervals of a full horizontal rotation.
This results in 18 images per scene and a total dataset size of 1656. 
Since the data collection procedure is highly sensitive to dynamic objects, additional manual verification is required. 
The final dataset consists of 1175 images, further separated into validation and testing splits of 400 and 775 images. 
We show some representative testing images in \fig{data:syns} and the distribution of images categories per split in \tbl{data:syns_cat}

\subsection{Training procedure} \label{sec:meth:train}
The first edition of \ac{mdec} focused on evaluating the \acl{sota} in self-supervised monocular depth estimation.
This included methods complemented by hand-crafted proxy depth maps or synthetic data. 
We expected most methods to be trained on Kitti~\cite{Geiger2013} due to its widespread use. 
However, we placed no restrictions on the training dataset (excluding SYNS/\acl{syns}) and encouraged participants to use additional training sources. 

To aid participants and give a strong entry point, we provided a public starting kit on GitHub\footnote{\codelink}.
This repository contained the training and evaluating code for 16 recent \ac{sota} contributions to \ac{mde}.
The baseline submission was the top F-Score performer out of all \ac{sota} approaches in this starting kit~\cite{Garg2016,Spencer2022}.
This consisted of a ConvNeXt~\cite{Liu2022} backbone and DispNet~\cite{Mayer2016} decoder.
The model was trained on the \acl{kez} split with an image resolution of \shape{192}{640}{}{} using only stereo view synthesis, the vanilla photometric loss and edge-aware smoothness regularization. 

\subsection{Evaluation procedure} \label{sec:meth:eval}
Participants provided their unscaled disparity predictions at the training image resolution. 
Our evaluation script bilinearly upsampled the predictions to the full image resolution and applied median scaling to align the predicted and ground-truth depths.
Finally, the prediction and ground-truth were clamped to a maximum depth of 100m. 
We omit test-time stereo blending~\cite{Godard2017} and border cropping~\cite{Eigen2015}.

%
\begin{figure}[htbp]
\centering
\input{Figures/*}
\label{fig:[}
\end{figure}
!t]{res_depth_viz}{res:depth_viz}

\subsection{Performance metrics} \label{sec:meth:metrics}
The predictions were evaluated using a wide variety of image/pointcloud/edge-based metrics. 
Submissions were ranked based on the F-Score performance~\cite{Ornek2022}, as this targets the structural quality of the reconstructed pointcloud.
We provide the units of each metric, as well as an indication if lower (\down) or higher (\up) is better.

\subsubsection{Image-based}

\heading{MAE}
Absolute error (m\down) as 
\begin{equation}
    \mae,
\end{equation}
where \acs{depth-gt} represents the ground-truth depth at a single image pixel~\acs{pix} and \acs{depth-pred} is the predicted depth at that pixel.

\heading{RMSE}
Absolute error (m\down) with higher outlier weight as
\begin{equation}
    \rmse.
\end{equation}

\heading{AbsRel}
Range-invariant relative error (\%\down) as 
\begin{equation}
    \absrel.
\end{equation}

\subsubsection{Pointcloud-based}

\heading{F-Score}
Reconstruction accuracy (\%\up) given by the harmonic mean of \textbf{P}recision and \textbf{R}ecall as 
\begin{equation}
    \fscore,
\end{equation}

\heading{Precision}
Percentage (\%\up) of predicted \ndim{3} points~\acs{point-pred} within a threshold~\acs{thr-3d} of a ground-truth point~\acs{point-gt} as 
\begin{equation}
    \precision,
\end{equation}
where $\myivr{\cdot}$ represents the Iverson brackets.

\heading{Recall}
Percentage (\%\up) of ground-truth \ndim{3} points within a threshold of a predicted point as 
\begin{equation}
    \recall.
\end{equation}
Following \"Ornek~\etal~\cite{Ornek2022}, the threshold for a correctly reconstructed point is set to 10 cm \ie $\acs{thr-3d} = 0.1$.
Note that Precision and Recall are only used to compute the F-Score and are not reported in the challenge leaderboard.

\subsubsection{Edge-based}

\heading{F-Score}

Pointcloud reconstruction accuracy (\%\up) computed only at ground-truth~\acs{edges-gt} and predicted~\acs{edges-pred} depth boundaries, represented by binary masks.

\heading{Accuracy}
Distance (px\down) from each predicted depth boundary to the closest ground-truth boundary as 
\begin{equation}
    \edgeacc,
\end{equation}
where \textit{EDT} represents the Euclidean Distance Transform.

\heading{Completeness}
Distance (px\down) from each ground-truth depth boundary to the closest predicted boundary as 
\begin{equation}
    \edgecomp.
\end{equation}
These metrics were proposed as part of the IBims-1~\cite{Koch2018} benchmark, which features dense indoor depth maps.

%
\begin{table}[htbp]
\scriptsize
\renewcommand{\arraystretch}{1.5}
\centering
\input{Tables/*}
\label{tbl:[}
\end{table}
!t]{res_metrics}{res:metrics}

%
\begin{figure}[!t]
\centering
\input{Figures/res_pts_viz}
\label{fig:res:pts_viz}
\end{figure}


\section{Challenge Submissions} \label{sec:submit}

\subsection*{Baseline}
\emph{%
\begin{tabular}{p{2cm}p{5cm}}
    J.\ Spencer\footnotemark[1] & j.spencermartin@surrey.ac.uk \\
    C.\ Russell\footnotemark[3] & cmruss@amazon.de \\
   S.\ Hadfield\footnotemark[1] & s.hadfield@surrey.ac.uk \\
     R.\ Bowden\footnotemark[1] & r.bowden@surrey.ac.uk \\
\end{tabular}
}

\noindent
Challenge organizers submission. 
Re-implementation of Garg~\cite{Garg2016} from the updated monocular depth benchmark\cite{Spencer2022}.
Trained with stereo photometric supervision with edge-aware smoothness regularization. 
Network is composed of a ConvNeXt-B backbone~\cite{Liu2022} with a DispNet~\cite{Mayer2016} decoder.
Trained for 30 epochs on \acl{kez} with an image resolution of \shape{192}{640}{}{}.

\subsection{Team 1 - \teamone}
\emph{%
\begin{tabular}{p{2cm}p{5cm}}
       H.\ Wang\footnotemark[6] & hwscut@126.com \\
      Y.\ Zhang\footnotemark[6] & yusheng.z1995@gmail.com \\
       H.\ Cong\footnotemark[6] & congheng@outlook.com \\
\end{tabular}
}

\noindent
Based on a ConvNext-B~\cite{Liu2022} with an HRDepth~\cite{Lyu2021} decoder. 
Trained with monocular and stereo data, along with proxy depth hints~\cite{Watson2019}.
This submission uses a large combination of losses, including the photometric loss with an explainability mask~\cite{Zhou2017}, autoencoder feature-based reconstruction~\cite{Shu2020}, virtual stereo~\cite{Godard2017}, proxy depth regression, edge-aware disparity smoothness~\cite{Godard2017}, feature smoothness~\cite{Shu2020}, occlusion regularization~\cite{Rui2018} and explainability mask regularization~\cite{Zhou2017}. 
The models were trained on \ac{kez} without depth hints and \ac{keb} with depth hints for 5 epochs and an image resolution of \shape{192}{640}{}{}.

\subsection{Team 2 - \teamtwo}
\emph{%
\begin{tabular}{p{2cm}p{5cm}}
   Z.\ K.\ Suri\footnotemark[8] & z.suri@eu.denso.com \\
\end{tabular}
}

\noindent
The depth and pose estimation networks used ConvNeXt-B~\cite{Liu2022} as the encoder, with the depth network complemented by a DiffNet~\cite{Zhou2021} decoder. 
Trained with both stereo and monocular inputs, using edge-aware regularization~\cite{Godard2017} and the min reconstruction photometric loss with automasking~\cite{Godard2019}.
A strong pose network is essential for accurate monocular depth estimation. 
This submission introduced a stereo pose regression loss.
The pose estimation network was additionally given a stereo image pair and supervised \wrt the know ground-truth camera baseline between them. 
The networks were trained on \acl{kez} with an image resolution of \shape{192}{640}{}{}.

\subsection{Team 3 - \teamthree}

\noindent
The author of this submission did not provide any details.

\subsection{Team 4 - \teamfour}
\emph{%
\begin{tabular}{p{2cm}p{5cm}}
       C.\ Zhao\footnotemark[9] & y20180082@mail.ecust.edu.cn \\
      M.\ Poggi\footnotemark[7] & m.poggi@unibo.it \\
       F.\ Tosi\footnotemark[7] & fabio.tosi5@unibo.it \\
      Y.\ Zhang\footnotemark[7] & youmin.zhang2@unibo.it \\
       Y.\ Tang\footnotemark[9] & yangtang@ecust.edu.cn \\
  S.\ Mattoccia\footnotemark[7] & stefano.mattoccia@unibo.it \\
\end{tabular}
}

\noindent
Trained on \ac{ke} with an image resolution of \shape{320}{1024}{}{}.
The depth network used the MonoViT~\cite{Zhao2022} architecture, combining convolutional and MPViT~\cite{Youngwan2022} encoder blocks. 
The network was trained using stereo and monocular support frames, based on the minimum photometric loss~\cite{Godard2019}, edge-aware smoothness~\cite{Godard2017} and \pnorm{1} proxy depth regression. 
Proxy depth labels were obtained by training a self-supervised stereo network~\cite{Lipson2021,Teed2020} on the Multiscopic dataset~\cite{Yuan2021}.
This dataset provides three horizontally aligned images, allowing the network to compensate for occlusions. 
The pretrained stereo network was trained using Center and Right pairs, but used the full triplet when computing the per-pixel minimum photometric loss. 
It was trained for 1000 epochs using \shape{256}{480}{}{} crops.

\section{Results} \label{sec:res}
\tbl{res:metrics} show the performance of the participants' submissions on the \acl{syns} test set.
As seen, most submissions outperformed the baseline in traditional image-based metrics (MAE, RMSE, AbsRel) across all scene types.
However, the baseline still achieved the best performance in both pointcloud reconstruction metrics (F-Score (Edges)). 
We believe this is due to the fact that most existing benchmarks report only image-based metrics. 
As such, novel contributions typically focus on improving performance on only these metrics.
However, we believe pointcloud-based reconstruction metrics~\cite{Ornek2022} are crucial to report, as they reflect the true objective of monocular depth estimation. 

As expected, all approaches transfer best to other Outdoor Urban environments, while the previously unseen Natural and Agriculture category provided a more difficult challenge. 
In most outdoor environments the baseline provides the best F-Score performance, while \teamone \& \teamfour improve on image-based metrics ($>0.5$ meter improvement in Outdoor Natural MAE). 
It is also interesting to note that all approaches improve the accuracy of the detected edges by roughly 15\%. 
Meanwhile, edge completeness is drastically reduced, implying that participant submissions are more accurate at extracting strong edges, but oversmooth predictions in highly textured regions. 
Finally, it is worth noting that the increased metric performance in indoor environments is likely due to the significantly decreased depth range. 

We show qualitative visualizations for the predicted depth maps and pointclouds in Figures \ref{fig:res:depth_viz} \& \ref{fig:res:pts_viz}, respectively.
The target images were selected prior to evaluation to reflect the wide variety of available environments.
Generally, we find that most predictions are oversmoothed and lack high-frequency detail. 
For instance, many models fill in gaps between thin objects, such as railings (second image) or branches (third image). 
As is expected, all submissions tend to perform better in urban settings, as they are more similar to the training distribution. 
The submission by \teamfour generally produces the highest-quality visualizations, with more detailed thin structures and sharper boundaries. 
This is reflected by the improved image-based metrics. 
However, as seen in the pointcloud visualizations in \fig{res:pts_viz}, these predictions still suffer from boundary interpolation artefacts that are not obvious in the depth map visualizations. 
This reinforces the need for more detailed metrics in these complex image regions. 

\section{Conclusions \& Future Work} \label{sec:conclusion}
This paper has presented the results for the first edition of \acl{mdec}. 
It was interesting to note that, while most submissions outperformed the baseline in traditional image-based metrics (MAE, RMSE, AbsRel), they did not improved pointcloud F-Score reconstruction. 
As expected, \acl{syns} represents a challenging dataset for current monocular depth estimation systems. 
We believe this to be due to the over-reliance on automotive training data.
Despite its availability and ease of collection, it does not contain varied enough scenarios to generalize to more complex natural scenes.
As such, it is likely that additional sources of training data are required to develop truly generic perception systems. 

Future editions of \ac{mdec} may expand to additionally evaluate supervised \ac{mde} approaches. 
This would help compare the \ac{sota} in both branches of research and help to determine the reliability of supervised networks.
We hope this provides a valuable contribution to the community and strongly encourage authors in this field to participate in future editions of the challenge.

\subsection*{Acknowledgements}
This work was partially funded by the EPSRC under grant agreements EP/S016317/1, EP/S016368/1, EP/S016260/1, EP/S035761/1.

{\small
\bibliographystyle{ieee_fullname}
\bibliography{main.bib}
}

\end{document}

%% file: preamble.tex
\usepackage{acronym}
\usepackage{xspace}
\usepackage{stmaryrd}  
\usepackage{booktabs}
\usepackage{tabularx}
\usepackage{multirow}
\usepackage{subfig}
\usepackage[font=footnotesize]{caption}
\usepackage{stackengine}
\usepackage{suffix}
\usepackage{adjustbox}
\usepackage{svg}
\usepackage{xfrac}
\usepackage{rotating}
\usepackage{xr}
\usepackage{pifont}  

\usepackage[textsize=scriptsize]{todonotes}
\presetkeys{todonotes}{inline}{}

\definecolor{ForestGreen}{HTML}{228b22}
\definecolor{Red}{HTML}{ff0000}  

\makeatletter
\newcommand*{\addFileDependency}[1]{
  \typeout{(#1)}
  \@addtofilelist{#1}
  \IfFileExists{#1}{}{\typeout{No file #1.}}
}
\makeatother

\DeclareRobustCommand{\nbd}{\nobreakdash-} 

\newcommand{\ndim}[1]{#1{\nbd}D}
\newcommand{\pnorm}[1]{L\ensuremath{_#1}}

\newcommand{\down}{\textcolor{black}{\ensuremath{\boldsymbol{\downarrow}}}}
\newcommand{\up}{\textcolor{black}{\ensuremath{\boldsymbol{\uparrow}}}}

\newcommand{\heading}[1]{\noindent \textbf{\small #1.}}

\newcommand{\mycaption}[2]{\caption[#1]{\textbf{#1.} #2}}

\newcommand{\addfig}[3][htbp]{%
\begin{figure}[#1]
\centering
\input{Figures/#2}
\label{fig:#3}
\end{figure}
}

\WithSuffix\newcommand\addfig*[3][htbp]{%
\begin{figure*}[#1]
\centering
\input{Figures/#2}
\label{fig:#3}
\end{figure*}
}

\newcommand{\addtbl}[3][htbp]{%
\begin{table}[#1]
\scriptsize
\renewcommand{\arraystretch}{1.5}
\centering
\input{Tables/#2}
\label{tbl:#3}
\end{table}
}

\WithSuffix\newcommand\addtbl*[3][htbp]{%
\begin{table*}[#1]
\scriptsize
\renewcommand{\arraystretch}{1.5}
\centering
\input{Tables/#2}
\label{tbl:#3}
\end{table*}
}

\def\eg{\emph{e.g}.~} 
\def\ie{\emph{i.e}.~}

\def\wrt{w.r.t.~} 
\def\etal{\emph{et al}.\xspace}

\newcommand{\fig}[1]{Figure~\ref{fig:#1}}
\newcommand{\tbl}[1]{Table~\ref{tbl:#1}}

\newcommand{\acromath}[3]{\acrodef{#1}[\(#2\)]{#3}} 
\newcommand{\myac}[1]{\text{\acs{#1}}}  

\DeclareMathOperator*{\mymin}{min}

\newcommand{\minus}{\scalebox{0.5}[0.75]{\ensuremath{-}}}

\newcommand{\shape}[4]{\ensuremath{  
#1 \times #2
\ifthenelse {\equal{#3}{}} {} {\times #3}
\ifthenelse {\equal{#4}{}} {} {\times #4}
}}

\newcommand{\sca}[1]{\ensuremath{#1}}  						   
\newcommand{\vct}[1]{\ensuremath{\textbf{\MakeLowercase{#1}}}} 
\newcommand{\mat}[1]{\ensuremath{\textbf{\MakeUppercase{#1}}}} 
\newcommand{\set}[1]{\ensuremath{\MakeUppercase{#1}}} 	       

\newcommand{\thr}{\delta}

\newcommand{\brackets}[3]{\ensuremath{\left#1 #2 \right#3}}
\newcommand{\mypar}[1]{\brackets{(}{#1}{)}}

\newcommand{\mynrm}[1]{\brackets{\|}{#1}{\|}}
\newcommand{\mymag}[1]{\brackets{|}{#1}{|}}  

\newcommand{\myivr}[1]{\brackets{\llbracket}{#1}{\rrbracket}}  

\newcommand{\easyfunc}[2]{\ensuremath{#1\mypar{#2}}}                 


%% file: acronyms.tex
\acrodef{syns}  [SYP] {SYNS{\nbd}Patches}
\acrodef{ke}    [KE]  {Kitti Eigen}
\acrodef{kb}    [KB]  {Kitti Benchmark}
\acrodef{kez}   [KEZ] {Kitti Eigen{\nbd}Zhou}
\acrodef{keb}   [KEB] {Kitti Eigen{\nbd}Benchmark}

\acrodef{vo}    [VO]    {Visual Odometry}
\acrodef{slam}  [SLAM]  {Simultaneous Localization and Mapping}
\acrodef{lidar} [LiDAR] {Light Detection and Ranging}
\acrodef{sota}  [SotA]  {State-of-the-Art}
\acrodef{mde}   [MDE]   {Monocular Depth Estimation}
\acrodef{mdec}  [MDEC]   {Monocular Depth Estimation Challenge}

%% file: symbols.tex
\acromath{pix}        { \vct{p} } {}

\acromath{point-gt}   { \vct{q} } {}
\acromath{point-pred} { \hat{\myac{point-gt}} } {}
 
\acromath{depth-gt}   { \sca{y} } {}
\acromath{depth-pred} { \hat{\myac{depth-gt}} } {}
 
\acromath{cloud-gt}   { \set{Q} } {}
\acromath{cloud-pred} { \hat{\myac{cloud-gt}} } {}
 
\acromath{edges-gt}   { \mat{M} } {}
\acromath{edges-pred} { \hat{\acs{edges-gt}} } {}
 
\acromath{thr-3d}     { \sca{\thr} } {}

\def\mae{\ensuremath{
    \sum \mymag{\acs{depth-pred} \minus \acs{depth-gt}}
}}
        
\def\rmse{\ensuremath{
    \sqrt{\sum \mypar{\acs{depth-pred} \minus \acs{depth-gt}}^2}
}}

\def\absrel{\ensuremath{
    \sum \frac{\mymag{\acs{depth-pred} \minus \acs{depth-gt}}}{\acs{depth-gt}}
}}

\def\precision{\ensuremath{
    \displaystyle 
    \sum_{\myac{point-pred} \in \myac{cloud-pred}} 
        \myivr
        { 
            \mymin_{\myac{point-gt} \in \myac{cloud-gt}}
            \mynrm{ \myac{point-gt} - \myac{point-pred} } 
            < 
            \acs{thr-3d} 
        }
}}

\def\recall{\ensuremath{
    \displaystyle
    \sum_{\myac{point-gt} \in \myac{cloud-gt}} 
        \myivr
        { 
            \mymin_{\myac{point-pred} \in \myac{cloud-pred}}
            \mynrm{ \myac{point-gt} - \myac{point-pred} } 
            < 
            \acs{thr-3d}
        }
}}

\def\fscore{\ensuremath{
    2 \cdot \frac{P \cdot R}{P + R}
}}

\def\edgeacc{\ensuremath{
    \displaystyle
    \sum \easyfunc{\textit{EDT}}{ \easyfunc{\acs{edges-pred}}{\acs{pix}} } \easyfunc{\acs{edges-gt}}{\acs{pix}}
}}

\def\edgecomp{\ensuremath{
    \displaystyle
    \sum \easyfunc{\textit{EDT}}{ \easyfunc{\acs{edges-gt}}{\acs{pix}} } \easyfunc{\acs{edges-pred}}{\acs{pix}}
}}

%% file: Figures/res_pts_viz.tex

\scriptsize

\def\first{0127}
\def\nums{0715}
\def\dataset{SynsPts}

\def\methods{
    gt/GT
    ,jspenmar/Baseline
    ,hwscut/\teamone
    ,zsuri/\teamtwo
    ,yiyiyihahaha/\teamthree
    ,Team/\teamfour
}

\foreach \method/\label in \methods{%
    \begin{tikzpicture}[every node/.style={inner sep=0,outer sep=0,anchor=south west}]
        \node (image) at (0, 0) {\includegraphics[width=0.45\linewidth]{\dataset/\method_\first.png}};
        \begin{scope}[x={(image.south east)},y={(image.north west)}]
            \draw (0.01, 0.01) node {\color{white}\textbf{\label}};
        \end{scope}
    \end{tikzpicture}
    \foreach \num in \nums{%
        \includegraphics[width=0.45\linewidth]{\dataset/\method_\num.png}
    } \\
}

\mycaption{\acl{syns} Pointcloud Visualization}{%
Converting the depth maps into poinclouds allows us to evaluate the quality of the reconstructed scene. 
All approaches can reliably estimate the road surface ground plane. 
However, object boundaries exhibit smooth interpolation  artefacts connecting them to background structures. 
Adapting to previously unseen indoor environments is still highly challenging. 
}